\documentclass{IEEEtaes}
\usepackage{colortbl}
\usepackage{color,array,amsthm}
\usepackage{graphicx}
\usepackage{amsmath}
\usepackage{float}
\usepackage{makecell}
\usepackage{longtable}
\usepackage{stfloats}
\usepackage{multirow}
\usepackage{graphics}
\usepackage{bm}
\usepackage{subfigure}
\usepackage[numbers,sort&compress]{natbib}
\usepackage{balance}
\usepackage[inkscapelatex=false]{svg}

\usepackage{booktabs}

\begin{document}

\title{ProbRadarM3F: mmWave Radar-based Human Skeletal Pose Estimation with Probability Map Guided Multi-Format Feature Fusion}

\author{BING ZHU}
\member{Senior Member,~IEEE}
\author{ZIXIN HE}
\author{WEIYI XIONG}
\author{GUANHUA DING}
\affil{Beihang University, Beijing, China} 


\author{TAO HUANG}
\member{Senior Member,~IEEE}
\affil{James CookUniversity, Cairns, Australia}

\author{WEI XIANG}
\member{Senior Member,~IEEE\\ La Trobe University, Melbourne, Australia}



\corresp{The work of Bing Zhu, Zixin He, and Weiyi Xiong was supported by National Natural Science Foundation of China under grant 62073015.}
\editor{Bing Zhu and Zixin He contribute equally and are co-first authors.\\This paper has been accepted by IEEE Transactions on Aerospace and Electronic Systems.\\Digital Object Identifier 10.1109/TAES.2025.3594328 }
\authoraddress{
Author’s addresses:
Bing Zhu, Zixin He and Weiyi Xiong are with School of Automation Science and Electrical Engineering, Beihang University, Beijing 100191, P.R.~China (e-mail: \href{mailto:zhubing@buaa.edu.cn}{zhubing@buaa.edu.cn};\href{mailto:hezixin2001@buaa.edu.cn}{hezixin2001@buaa.edu.cn};\href{mailto:weiyixiong@buaa.edu.cn}{weiyixiong@buaa.edu.cn}). \\
Guanhua Ding is with the School of Electronics and Information Engineering, Beihang University, Beijing 100191, P.R.~China (e-mail: \href{mailto:buaadgh@buaa.edu.cn}{buaadgh@buaa.edu.cn}). \\
Tao Huang is with College of Science and Engineering, James Cook University, Cairns, Australia (e-mail: \href{mailto:tao.huang1@jcu.edu.au}{tao.huang1@jcu.edu.au}).\\
Wei Xiang is with the School of Computing, Engineering and Mathematical Sciences, La Trobe University, Melbourne, Australia (e-mail: \href{mailto:w.xiang@latrobe.edu.au}{w.xiang@latrobe.edu.au}).  {\itshape (Corresponding author:Bing Zhu)}
\thanks{}
}

\markboth{ZHU ET AL.}{PROBRADARM3F}
\maketitle

\begin{abstract}
Millimeter wave (mmWave) radar is a non-intrusive, privacy-preserving, and cost-effective device, shown to be a viable alternative to RGB cameras for indoor human pose estimation. However, the challenge lies in fully leveraging the reflected radar signals for accurate pose estimation.
To address this major challenge, this paper introduces a probability map guided multi-format feature fusion model, ProbRadarM3F. This is a radar feature extraction framework using a traditional FFT method in parallel with a probability map based positional encoding method. ProbRadarM3F fuses the traditional heatmap features and the positional features, then effectively achieves the estimation of $14$ keypoints of the human body. Experimental evaluation on the HuPR dataset proves the effectiveness of $69.9\%$ in average precision (AP). The emphasis of our study is on utilizing position information in radar signals for estimating human skeletal pose. This provides direction  
for investigating other potential non-redundant information from mmWave radar.
\end{abstract}


\begin{figure}[b]
\centering
\includegraphics[scale=0.25]{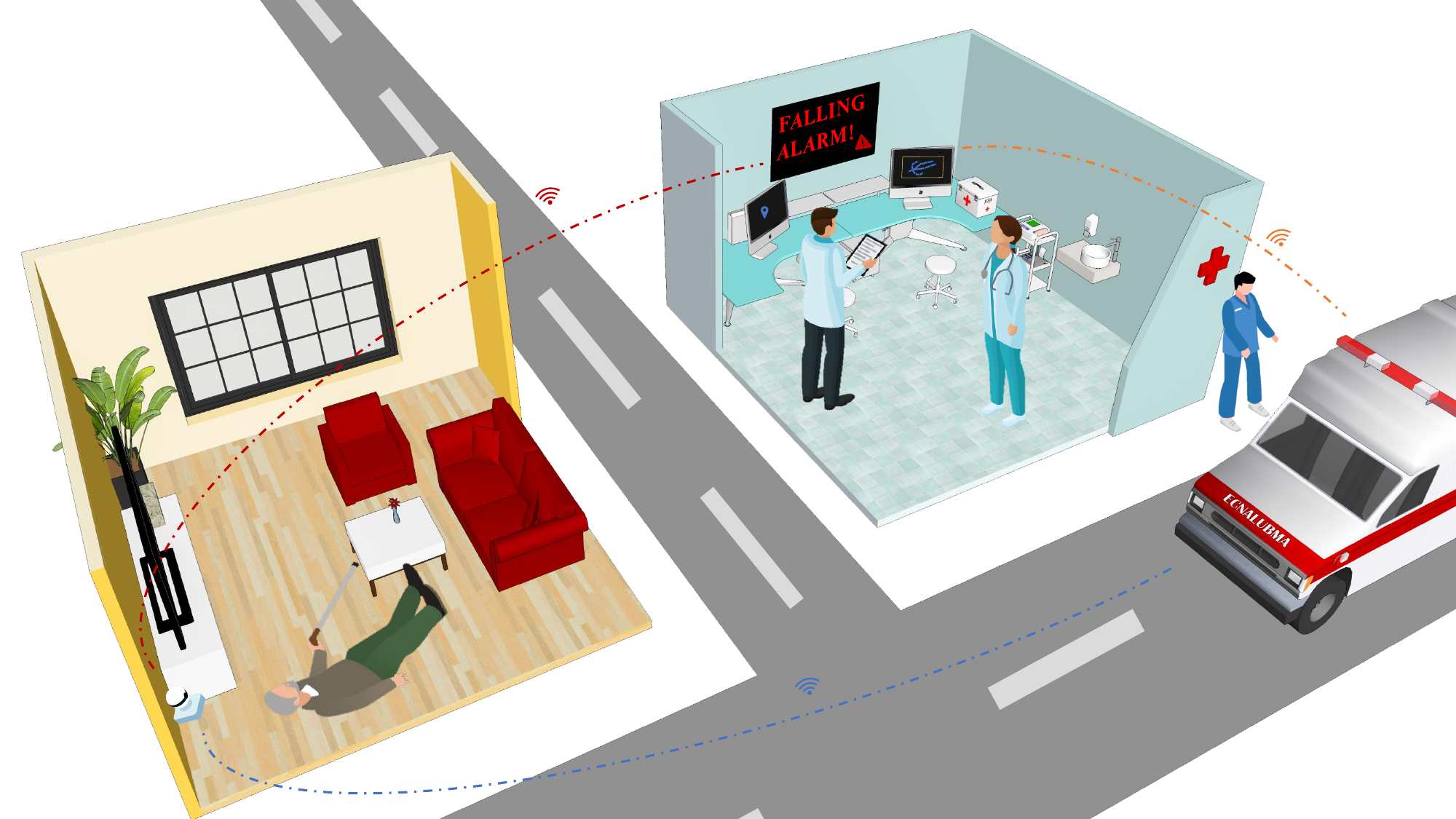}
\caption{
\raggedright
Illustration of a medical IoT application: if someone receiving care at home falls accidentally, it can be immediately detected, triggering an alarm at a remote care center and prompting the automatic dispatch of an ambulance. Our research focuses on utilizing mmWave radar to estimate the human skeletal pose, a critical sensor component in this example of a medical IoT application. This approach preserves the user's privacy while still providing the necessary monitoring information to the care center.}
\label{fig_iot}
\end{figure}

\section{INTRODUCTION}
S{\scshape mart} medical care is paramount in hospitals and nursing homes in today's society. 
To improve healthcare outcomes and ensure patient safety, there is an increasing need for human pose monitoring. The Internet of Things (IoT) has become instrumental in this regard, providing continuous and efficient monitoring capabilities \cite{HAR_IoT,MC_IoT}. 
As shown in Fig. \ref{fig_iot}, IoT with human activity monitoring is essential to improve the convenience, safety, and interactivity of medical care. 
Traditionally, monitoring systems have relied on RGB cameras to estimate human pose and track human activity \cite{HPESurvey,RGBHPEsurvey}. 
However, RGB-based solutions can be unreliable in low-light or occluded settings. More critically, cameras raise substantial privacy concerns in sensitive indoor environments such as nursing homes, bedrooms, and hospital wards. People are reluctant to accept devices that capture visual imagery because of the risk of data leakage. Consequently, using indoor cameras to monitor the pose of people who are in need, like the elderly or patients, is often considered unacceptable.

Due to these mentioned reasons, recent advancements focus on reducing privacy concerns by using non-visual signal-based sensing methods \cite{WifiSurvey,RadarSurvey}. 
These signals contain no human facial information or visual photographs, thus protecting human privacy in IoT application scenarios \cite{Virteach, MetaFi}. 
Millimeter wave (mmWave) radar has emerged as a prominent technology due to its high resolution and cost-effectiveness. 
MmWave radar can accurately estimate human skeletal poses without capturing identifiable visual information, making it ideal for preserving privacy. 
This paper focuses on mmWave radar-based human skeletal pose estimation, a critical component in smart sensor systems for medical care and smart home IoT applications.

MmWave radar, a robust yet low-cost sensor, has been applied for environment perception in advanced driver-assistance systems (ADAS) \cite{xiong2022contrastive,liu2022deep,SMURF,LXL,RaLiBEV,EOT_vs_POT_4D_Radar_3D_MOT} and cooperative intelligent transportation systems (C-ITS) \cite{V2X_CP_Overview} in recent years. 
Considerable frontier research on mmWave radar has focused on achieving human activity sensing tasks. However, significant challenges remain.
In the classical radar processing pipeline, various levels of data representation exist. 
For human skeletal pose estimation, point clouds and heatmaps are the most widely used forms of data representation. 
However, point cloud generation relies on hand-crafted parameters, and the process itself loses Doppler information and some positional data, leading to difficulties in capturing finer joint features. 
Heatmaps are often generated using the Fast Fourier Transform (FFT) method \cite{Mmwavefundamentals}, which can extract range, Doppler, and angle of arrival (AoA) information from the raw radar signal to create different types of heatmaps. 
However, much of the information in the heatmap is considered redundant and not valued. 
Recently, many pose estimation works have been realized through a point cloud \(+\) neural network pipeline\cite{STMCN} or a heatmap \(+\) neural network pipeline. 
Some studies have focused on improving the deep learning capability of networks and algorithms to address the mmWave radar-based precision problem. 
However, the importance of capturing the information present in the radar signal itself is often neglected, leading to the loss of significant effective information.

To address the inefficiencies in traditional heatmap-based methods, a new method is proposed, constructing an efficient feature extraction framework.
The method incorporates two feature extraction branches, generating multi-format features, i.e., distinct feature sets derived from the same raw data using different extraction strategies: the frequency features from heatmaps, and the positional features guided by probability maps. Specifically, one branch adheres to conventional procedures, utilizing the FFT method to generate heat maps to extract range, Doppler, azimuth, and elevation information from mmWave radar signals. 
The other branch introduces a probability map generation and positional encoding method, designed to enhance the utilization of positional information in radar signals.
%
Our model fuses features in two formats (probability-guided positional features and traditional heat map features) and takes into account the influence and help of the frames before and after the target frame in the fusion.
The contributions of this work include:
\vspace{-5pt}
\begin{itemize}
\item[$\bullet$] 
We propose ProbRadarM3F, a mmWave radar-based multi-format feature fusion model built on the model in \cite{Hupr}, for the estimation of human skeletal pose. The model is designed to utilize raw mmWave radar signals, with different formats of features extracted from two branches: the FFT branch and the introduced ProbPE branch.
\item[$\bullet$]
The ProPE branch proposed in this paper is a novel approach to generate probability map-guided positional encoding, as an additional format of radar feature.
To the best of our knowledge, this study represents the first attempt to extract additional probabilistic and positional features from mmWave radar signals to aid in pose estimation. 
By fusing features in different formats from multiple frames, the keypoint prediction accuracy is significantly increased. 
Thus, ProbRadarM3F can serve as a new baseline for research and application in radar position information exploitation.
\item[$\bullet$] 
To validate the effectiveness of ProbRadarM3F, experiments were conducted on the same dataset as HuPR \cite{Hupr}, providing purely raw radar signals.
The results demonstrate that the design of a probability map-guided positional encoding strategy greatly improves the recognition performance, leading to a significant enhancement in the precision of human skeletal pose prediction.

\end{itemize}

This paper is organized as follows: Section \ref{sec related} provides a comprehensive review of the research and methodology of mmWave radar for indoor sensing applications, particularly for human skeletal pose prediction. 
Section \ref{sec methods} details the implementation of the proposed ProbRadarM3F model.                                                                                                                          
In Section \ref{sec results}, experiments conducted with ProbRadarM3F on the HuPR dataset are presented, and the results are evaluated. 
The advantages of its pose prediction capabilities are demonstrated. 
Finally, Section \ref{sec conclusion} summarizes the approach and findings and suggests potential directions for future research.

\begin{figure*}[t]
\centering
\includegraphics[width=\linewidth,scale=1.10]{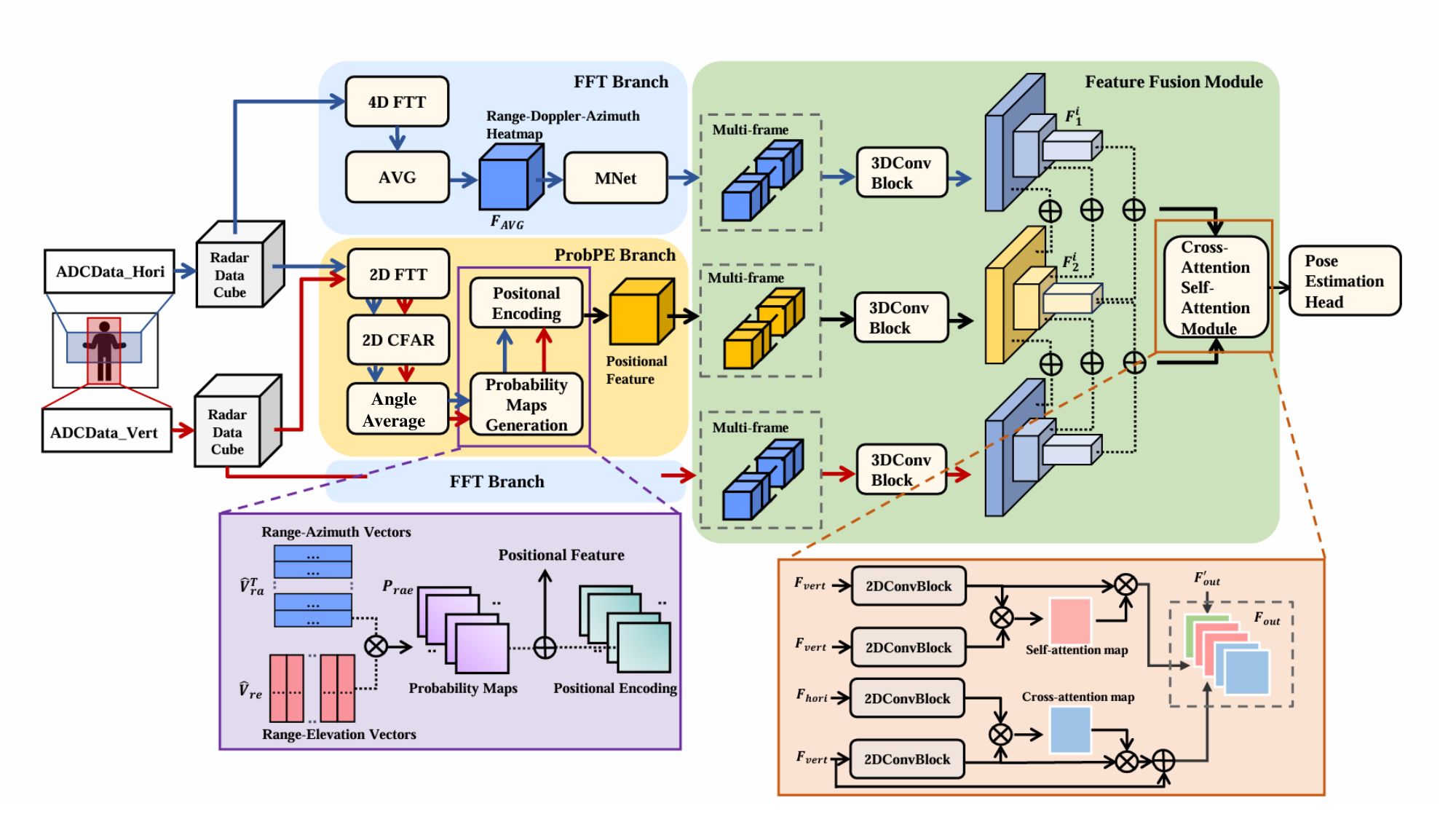}
\caption{
\raggedright
Illustration of our proposed ProbRadarM3F model. This model consists of two main branches, namely the FFT branch and the ProbPE branch. Different colored arrows represent data streams from different radars. The initial input is the raw ADC data from two vertically placed mmWave radars. After processing the raw radar data into a radar data cube, the FFT branch processes the data to extract features from range-Doppler-azimuth heatmaps and range-Doppler-elevation heatmaps based on 4D-FFT. The ProbPE branch applies positional encoding to extract features from generated radar probability maps. In the following module, the multi-frame and multi-format features are fused, and a cross- and self-attention module \cite{Hupr} is introduced to generate estimated skeletal pose.}
\label{fig1}
\end{figure*}

\section{RELATED WORK}\label{sec related}
\subsection{Millimeter wave Radar Indoor Sensing} 
The field of millimeter wave (mmWave) radar for indoor sensing has gained substantial advancements in recent years, driven by the growing demand for non-invasive, privacy-preserving technologies for human presence and activity detection. 
MmWave radar offers unique advantages for indoor sensing applications. 
The mmWave frequency can penetrate non-metallic objects while providing high-resolution imaging capabilities \cite{mmReview}. 
Its utility is increasingly evident in through-wall sensing \cite{throughWallCNN}, human detection and recognition \cite{mmwaveHumanSurvey}, and vital sign monitoring \cite{breathid_TAES,VitalSignalMonitor_IOTJ_1,VitalSignalMonitor_IOTJ_2,VitalSignalMonitor_ICASSP}.

Initially, researchers used radar signal processing to extract waveform features, combining them with target distance, orientation, altitude, and motion information to perform simple action classification\cite{mmwaveApplications}. 
Later, advanced technologies, e.g., deep learning, were introduced for more accurate indoor sensing projects based on basic mmWave radar data. 
For instance, Wang et al. proposed a sequence-to-sequence (seq2seq) 3D temporal convolutional network with a self-attention method to estimate human hand pose \cite{mmPose}. 
Yi et al. utilized a multi-person detection model based on long short-term memory (LSTM) to determine human presence and localize their positions using a single 60GHz mmWave radio \cite{Mmface}. 
Ding et al. developed a novel hybrid neural network model that explores multi-domain fusion of radar information by utilizing three kinds of 2D domain spectra. They combine 1D and 2D convolutional neural network and a recurrent neural network to capture rich features through multiple domains \cite{HPE_TAES}. 
Wu et al. pioneered the work of segmenting human silhouettes from millimeter wave RF signals by locating human reflections in radar frames and extracting features from surrounding signals with a human detection module and from aggregated frames with an attention-based mask generation module \cite{RFmask}. 

Although these studies have been highly successful, they have focused more on changes in the structure of deep learning networks and neglected how to extract more valid features from the radar signal. 
Given the inherent limitations of mmWave radar data, this oversight may result in the loss of valuable information during processing.
Therefore, in this paper, we pay special attention to the position information contained in the radar signals, which is long unnoticed but rich in  information. The positional feature is extracted and combined with traditional features to depict human body poses.

\subsection{MmWave based Human Skeletal Pose Estimation} 


Since the application demands both privacy and precision, the advent of mmWave radar technology has expanded the horizons of human pose estimation and recognition. 
A notable research direction in human pose estimation is the recognition of skeletal pose. 
The advantage of human skeletal pose estimation lies in its ability to provide a simplified but informative representation of human pose and movements, facilitating efficient and accurate recognition of human activity. 
Skeletal pose prediction derived from mmWave radar signals features critical spatial and temporal information about human activity without the privacy concerns associated with optical imaging.

RF-Pose \cite{ThroughWallHPE} is one of the pioneering works that first considered human skeleton reconstruction. 
Zhao et al. proposed a deep network from Frequency Modulated Continuous Wave(FMCW) radar signals to estimate coarse parts of the human skeleton under the supervision of visual information. 
They subsequently improved the model to successfully predict 14 human joints, including the head, neck, shoulders, elbows, wrists, hips, knees, and feet. 
RF-Pose and its follow-up work utilized expensive customized hardware systems. 
Nowadays, more studies are based on low-cost mass-produced industrial millimeter wave radar and have demonstrated good performance. 
Ding et al. presented a kinematic-constrained learning architecture that incorporates kinematic constraints with neural network learning for skeleton estimation based on range-Doppler heatmaps \cite{RadarKinematicLearning}. 
Kong et al. performed a convolution operation on the Range-Doppler Profile to detect the corresponding ranges and designed a two-stream deep learning architecture to extract body shape and motion features to predict skeleton joint coordinates and reconstruct body posture \cite{m3track}. 
Cao et al. developed a model that incorporates part-level range-Doppler maps for individual body parts with local kinematic constraints and global constraints for reconstructing the human skeleton \cite{GlobalLocalNetHPE}.
Cao et al. proposed a two-stage framework for task-specific feature purification, which initially employs an adversarial auto-encoder to extract interpersonal-independent features, followed by a feature disentanglement module to eliminate components irrelevant to the pose \cite{pureHPE_TAES}.

While these studies have laid an important foundation in the estimation of skeletal pose using mmWave radar, they focus primarily on high-cost hardware systems, require substantial visual information supervision, or rely on complex processing of heat maps and point clouds. Without exploring the integration of probabilistic and positional features in a unified framework, existing methods may limit their applicability to various practical scenarios.
In this work, our proposed extracting positional features based on probability maps are highly beneficial in accurately predicting human key joints positions and reconstructing human skeletal pose. This innovative approach enhances the robustness and precision of skeletal pose estimation.

\section{PROPOSED METHODS}\label{sec methods}


In this section, detailed descriptions of ProbRadarM3F for human skeletal pose estimation, built upon \cite{Hupr}, with additional fusion of multi-format features from mmWave radar data, are provided. 
The structure of the model is illustrated in Fig. \ref{fig1}. After processing the raw signal from mmWave radar, multi-format features are extracted from the radar data through the FFT and ProbPE branches. 
Multi-frame features in different formats from the two branches are fused.
Finally, the coordinates of human skeletal joints are predicted from the fused features via a cross-self-attention mechanism.

\subsection{FFT Branch: Radar Data Processing and Range-Doppler-Angle Map Generating}


The dataset used provides only unprocessed raw radar data acquired by the DAC1000 system. 
Therefore, the radar signal is pre-processed to convert the raw analogue-to-digital converter (ADC) data into a structured format, specifically a radar data cube, for efficient signal processing and analysis.


The raw ADC data, captured in binary format, is initially one-dimensional, representing interleaved ADC samples across multiple channels. 
It is structured by organizing the data into a two-dimensional matrix to separate the individual low voltage differential signaling (LVDS) channels. 
By segregating the samples from even and odd channels, the real and imaginary components of the complex signal are constructed, ensuring the retention of phase information necessary for subsequent processing tasks. 
Ultimately, the complex data is reformatted into a radar data cube to accurately reflect the sequence of radar data acquisition. 
This reorganization involves structuring the data into blocks corresponding to each chirp across multiple transmit (TX) and receive (RX) channels. 
The radar data cube accurately retains information about the spatial and temporal dimensions of the observed scene.


As shown in the FFT branch in Fig. \ref{fig1}, a comprehensive analysis is initiated by performing a 4-dimensional FFT (4D FFT) along the four axes: ADC samples, chirps, horizontal antennas, and vertical antennas as follows:
\begin{equation}
\begin{aligned}
F(h, i, j, k)= \sum_{n=0}^{N-1} \sum_{m=0}^{M-1} \sum_{p=0}^{P-1} \sum_{q=0}^{Q-1} f(n, m, p, q) \\
\times e^{-j2\pi\left(\frac{h n}{N} + \frac{i m}{M} + \frac{j p}{P} + \frac{k q}{Q}\right)},
\end{aligned}
\end{equation}
%
%
where ${f(n,m,p,q)}$ represents the original input signal across ADC samples ${n}$, chirps ${m}$, horizontal antennas ${p}$, and vertical antennas ${q}$, and ${F(h, i, j, k)}$ denotes the results of the Fourier transform. 
${N, M, P, Q}$ are the sizes of the dimensions, while ${h, i, j, k}$ are the indices in the frequency domain for the corresponding dimensions.
This process yields detailed range-Doppler-azimuth-elevation maps.
Additionally, a certain amount of chirps within a specific velocity range is uniformly sampled in the Doppler dimension to filter out irrelevant information in the radar signal. 
This strategy effectively eliminates extraneous signal components, thereby streamlining the dataset for enhanced processing efficiency.


The dataset utilized in this study provides data collected by two vertically placed radars. 
Due to the limited resolution and detection range in elevation angle directions of the adopted radars, the elevation information from each radar is averaged in the processing as follows:

\begin{equation}
\begin{aligned}
F_{\mathrm{avg}} = \frac{1}{Q} \sum_{k_q=0}^{Q-1} F(k_m,k_n,k_p,k_q).
\end{aligned}
\end{equation}


While the azimuth information of the horizontally placed radar is retained, the azimuth information of the vertically placed radar is considered as elevation angle information for the overall data. 
Eventually, the range-Doppler-azimuth map from the horizontal radar and the range-Doppler-elevation map from the vertical radar are obtained. 
Following the pioneering work \cite{Hupr}, to combine dynamic information across varying chirps within the range-Doppler-angle maps, the features from multiple chirps are merged by employing the M-Net \cite{RODNet}. 
The M-Net module utilizes a neural network to fuse different chirp information within a frame, instead of the traditional method, to solve for Doppler velocities, thereby outputting the merged features of the frame.


\subsection{ProbPE Branch: Probability Map Generating and Positional Encoding}


As shown in Fig. \ref{fig1}, in the ProbPE branch, the radar data cube is initially subjected to a two-dimensional FFT (2D FFT). 
The FFT is executed in the fast-time dimension (the dimension of digitized chirp samples) to obtain range information. 
Due to the significant impact of velocity frequency effects in the slow-time dimension, where multiple frames correspond to the same range unit, an FFT is also conducted in this dimension to obtain the Doppler frequency.

\begin{figure*}[htbp]
\centering
\includegraphics[width=\linewidth,scale=1.0]{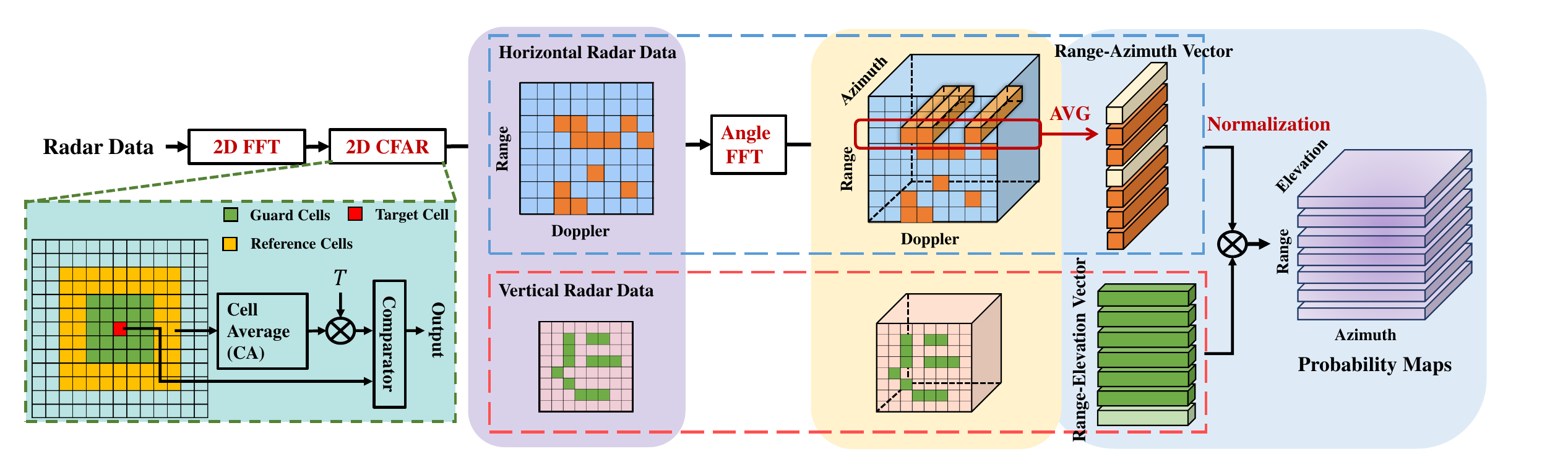}
\caption{
\raggedright
Illustration of the probability map generation, which is a  more detailed representation of the lower left part of Fig.2.
It starts by filtering out the range bin through 2D CFAR on the range-Doppler heatmaps. 
Angle information is then extracted from the selected range, producing range-azimuth and range-elevation vectors. 
These vectors are normalized, transposed, and multiplied to create probability maps, indicating the likelihood of the target appearing at a specific position. 
The probability map serves as a guide for the positional encoding method.
}
\label{fig2}
\end{figure*}

To minimize irrelevant information that increases computation or interferes with feature extraction, a constant false-alarm rate (CFAR) method \cite{RadarPrinciples} is employed. 
This method is crucial in distinguishing between targets and interference noise based on intensity differences. 
A two-dimensional CFAR (2D-CFAR) is applied to range-Doppler maps to select the elements containing more valid information. 
In the 2D-CFAR detection process, targets are detected as accurately as possible to reduce false alarms. 
Guard cells and reference cells are introduced near each detection cell, as shown in the lower left corner of Fig. \ref{fig2}. 
Cell Averaging CFAR (CA-CFAR) is a common variant of CFAR, which uses the following formula to calculate the threshold for each cell under examination:
\begin{equation}
\begin{aligned}
T = \alpha \cdot \frac{1}{N} \sum_{i=1}^{N} X_{i},
\end{aligned}
\end{equation}
%
%
where ${T}$ represents the threshold for the cell, ${X_{i}}$ represents the amplitude values of the reference cells surrounding the cell under test, ${N}$ represents the number of reference cells considered, and ${\alpha}$ is a scaling factor determined by the desired false alarm rate. 
If the intensity of the target surpasses that of interference noise, as ${I_\mathrm{target} > I_\mathrm{noise} \times T}$, it indicates the presence of the target.


Subsequently, probability maps are derived from the filtered range bins, which serve as a foundational step in understanding the spatial distribution of potential targets. 
The concatenation set for the filtered range bins from the two radars is taken separately. 
The phase differences among various antennas yield distinct spikes, indicating frequency components from different directions. 
The angle information is obtained by performing FFT on target data from distinct virtual antennas. 
Angle FFT is executed to derive azimuth from the horizontal radar and elevation from the vertical radar. 
An averaging calculation in the Doppler dimension is performed because the focus at this point is on probability and position information. 
For the non-overlapping range bins filtered out from two radars, the Doppler dimensions are averaged directly to fill the corresponding range-angle vectors, facilitating the subsequent multiplication operations.
Thus, range-azimuth and range-elevation vectors from the two radars are extracted, respectively.
These vectors contain information on the radar's intensity at specific ranges and angles, effectively characterizing the probability of a target's presence at a given location. 
Consequently, radar probability maps are constructed, as illustrated in Fig. \ref{fig2}.

The normalization of these vectors is performed to ensure uniformity in the data. 
Following normalization, the range-azimuth and range-elevation vectors at identical ranges are transposed and multiplied to generate range-azimuth-elevation probability maps:
\begin{equation}
\begin{aligned}
P_\mathrm{rae}(r, \theta, \phi) = \hat{V}_\mathrm{ra}(r, \theta)^T \cdot \hat{V}_\mathrm{re}(r, \phi),
\end{aligned}
\end{equation}
where ${P_\mathrm{rae}(r, \theta, \phi)}$ represents the probability map indicating the likelihood of detecting a target at range ${r}$, azimuth ${\theta}$, and elevation ${\phi}$. ${\hat{V}\mathrm{ra}(r, \theta)}$ represents the normalized range-azimuth vector, and ${\hat{V}\mathrm{re}(r, \phi)}$ denotes the normalized range-elevation vector.


These probability maps not only indicate the probability of detected elements but also contain significant geometric information, suggesting further exploitable position data. 
In advanced deep learning methodologies, the Transformer \cite{Transformer} enhances spatial perception through a sine coding formula for positional encoding.
DETR \cite{DERT} incorporates the Transformer model into target recognition, applying sine positional encoding to 2D images.


In our project, for each range $r$, the azimuth-elevation probability map can be treated as a two-dimensional representation. 
Therefore, sine positional encoding is introduced to extract positional information from the maps. 
The sine encoding formula is as follows:
\begin{equation}
\begin{aligned}
{PE}^r_{\left(\mathrm{{pos}}_{\theta}, 2 i\right)}=\sin \left( \mathrm{pos}_{\theta} / 10000^{\frac{2i}{d}}\right), 
\end{aligned}
\end{equation}

\begin{equation}
\begin{aligned}
{PE}^r_{\left(\mathrm{{pos}}_{\theta}, 2 i+1\right)}=\cos \left( \mathrm{pos}_{\theta} / 10000^{\frac{2i}{d}}\right), 
\end{aligned}
\end{equation}

\begin{equation}
\begin{aligned}
{PE}^r_{\left(\mathrm{pos}_{\phi}, 2 i\right)}=\sin \left( \mathrm{pos}_{\phi} / 10000^{\frac{2i}{d}}\right), 
\end{aligned}
\end{equation}

\begin{equation}
\begin{aligned}
{PE}^r_{\left(\mathrm{pos}_{\phi}, 2 i+1\right)}=\cos \left( \mathrm{pos}_{\phi} / 10000^{\frac{2i}{d}}\right),
\end{aligned}
\end{equation}
%
%
where $(\mathrm{pos}_{\theta}, \mathrm{pos}_{\phi})$ represents the computed position and \(d\) represents the dimension of the positional encoding vector. 
This formula intricately encodes position information by encoding each coordinate, \(\mathrm{pos}_{\theta}\) and \(\mathrm{pos}_{\phi}\), into a unique 32-dimensional vector through sine and cosine functions. 
This approach captures the essence of spatial variations across the probability maps.


By integrating positional encoding with the original probability features, the model is endowed with an enhanced capacity to recognize and interpret spatial relationships, significantly improving its performance on subsequent tasks. 
The features are then fed into the subsequent feature extraction module, which aligns with the structure of the other branch.

\subsection{Multi-format Feature Fusion and Estimation Head}


The features obtained from both branches are based on single-frame radar data. 
In human body movement, both previous and subsequent frames can be used as references for the current frame action information. 
To better utilize temporal information, the multi-frame information is jointly processed, considering the inherent temporal continuity of human motion. 
Data from multiple frames before and after the target frame are integrated, thus harnessing richer temporal features.


%
In our model, multi-format refers to the extraction and fusion of features from different representations of the radar data. Specifically, features from two branches utilize two distinct formats: the FFT branch captures frequency domain features from traditional heatmaps, while the ProbPE branch generates probability maps that highlight the positional features. By combining these features of various formats, our model leverages the complementary strengths of both branches.

To integrate multi-format spatio-temporal features, multiple 3D convolutional layers are used to aggregate information. 
These 3D convolutional blocks help extract and consolidate features across both spatial and temporal dimensions. 
For each spatial scale, a 3D convolution block aggregates the residual temporal information, yielding three-layer encoded features. 
In addressing the challenge of integrating multi-scale features from distinct processing branches, the output dimensions of encoded features across layers are normalized. 
This uniformity facilitates the direct summation of positional and probability features with features from the other branch. 
Unlike the conventional concatenation method, this direct addition method preserves spatial coherency, particularly for positional features. 
The fused features are obtained as follows:
\begin{equation}
\begin{aligned}
F_{f}^{i} = F_{1}^{i} + F_{2}^{i}\quad (i = 1, 2, 3),
\end{aligned}
\end{equation}
%
%
where \(i\) indicates the layers after 3D convolution. 
\(F_f\) represents the fused feature of each layer, and \(F_1\), \(F_2\) represent features from the two branches, respectively.


Following the pioneering work \cite{Hupr}, the layer features obtained by summing go through a cross- and self-attention decoding module, specifically designed to enhance human pose detection by leveraging contextual information across and within the frames captured by both horizontal and vertical radar systems. 
The core consists of multiple decoder layers, each designed to handle features at varying scales and complexities. 
At each scale, the network combines basic 2D blocks for initial feature transformation and flattens to get an attention map in each attention system. 
As shown in the lower right corner of Fig. \ref{fig1}, the cross-attention mechanism combines the same layer of features from both radars. 
In the cross-attention mechanism, the features from the horizontal radar are used as key and value, and the features from the vertical radar as queries. 
The roles of key, query, and value are then switched to perform the same operation. 
Since the horizontal and vertical radar capture features in different aspects, i.e., azimuth and elevation data, a residual connection is crucial to avoid directly correlating these features and to ensure training stability.
In addition, all keys, queries, and values in the self-attention system are from the same radar to enhance the internal structure of individual features. 
There are no skipped connections to generate self-participating residual features.


The predicted keypoint heatmaps are obtained through the attention decoding module from the multi-layer features. 
Knowing the location of certain keypoints can help in predicting the location of other keypoints. 
Therefore, same as \cite{Hupr}, the pose refinement module of the Graph Convolutional Network (GCN) is used to refine the keypoint heatmap \cite{Graphpcnn}. 
A 3-layer GCN is used to perform feature propagation and inference, refining the keypoint predictions using the mutual position information of the physical connections between keypoints. 
The generated new heatmap is used as a common reference to generate the final keypoint locations. 
There is a mismatch between the coordinate systems for radar signals and keypoint heatmaps, making it impossible to directly use keypoint coordinates to locate radar features.


To facilitate the end-to-end training, a strategy of imposing a pixel-wise binary cross-entropy loss on both the initial keypoint heatmaps and the GCN-refined keypoint heatmaps is employed. 
The objective function is formulated by
\begin{equation}
\begin{aligned}
L = L_\mathrm{bce}(\hat{H}, G) +   L_\mathrm{bce}(H, G),
\end{aligned}
\end{equation}
%
%
where ${H}$ and ${\hat{H}}$ represent the initial keypoint heatmaps and refined keypoint heatmaps, respectively, and ${G}$ represents the generated ground truth keypoint heatmaps based on Gaussian distribution.
To illustrate, the binary cross-entropy loss \(L_\mathrm{bce}(H, G)\) is defined by 
\begin{equation}
\begin{aligned}
L_\mathrm{bce}(H, G) = -\sum_{c,i,j} G_{c,i,j} \log(H_{c,i,j}) \\+ (1 - G_{c,i,j}) \log(1 - H _{c,i,j}),
\end{aligned}
\end{equation}
where \(c,i,j\) are derived from the channel, width and height of the joint prediction heatmaps \(H\) and groud truth heatmaps \(G\), which follow a Gaussian distribution.

\section{EXPERIMENTS AND RESULTS}\label{sec results}

\begin{figure*}[!t]
\centering
\includegraphics[width=\linewidth,scale=1.00]{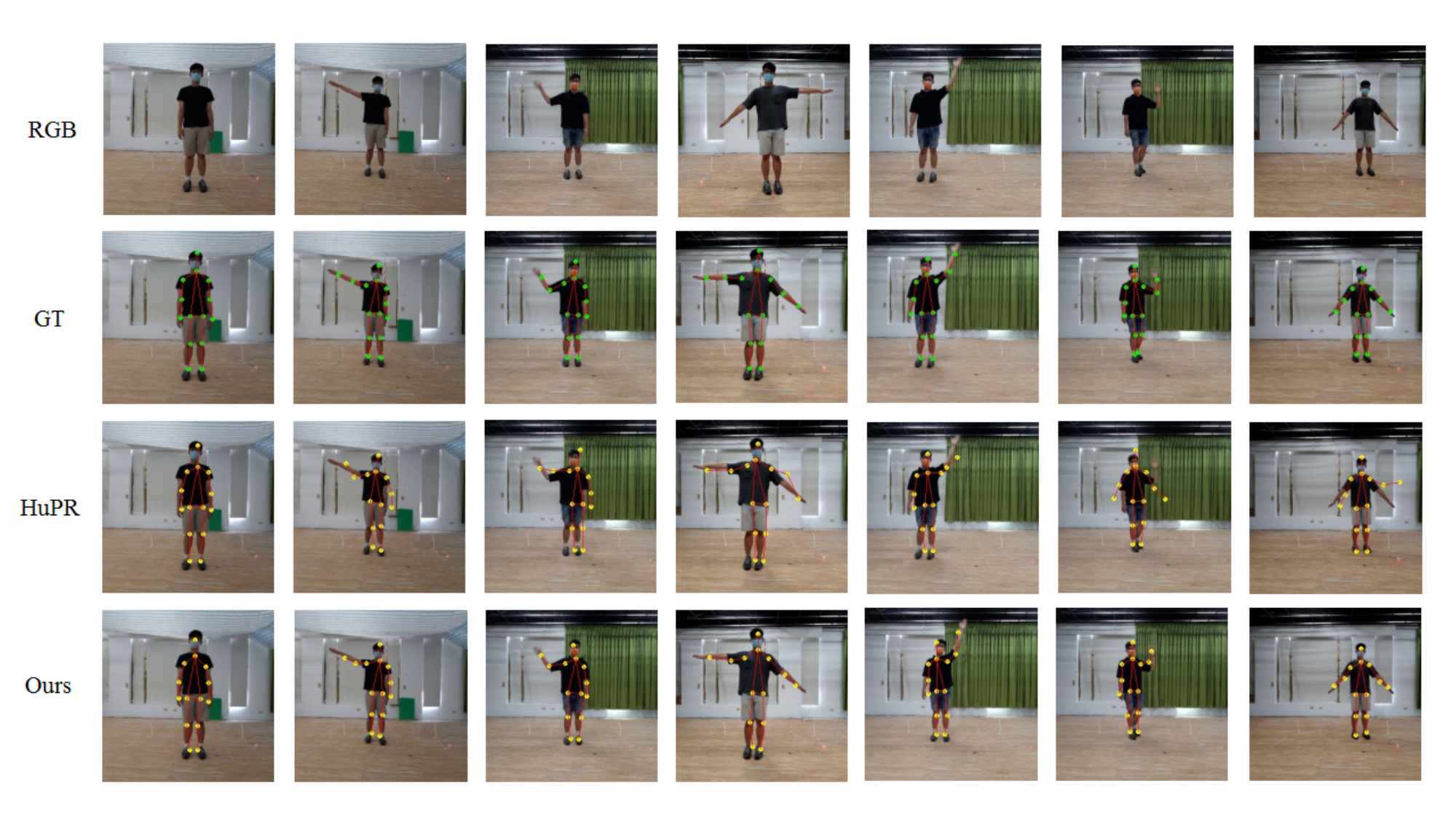}
\caption{Visualisation of comparison in state-of-the-art approaches and our proposed model.}
\label{fig3}
\end{figure*}

\begin{figure}[!t]
\centering
\includegraphics[width=\linewidth,scale=1.00]{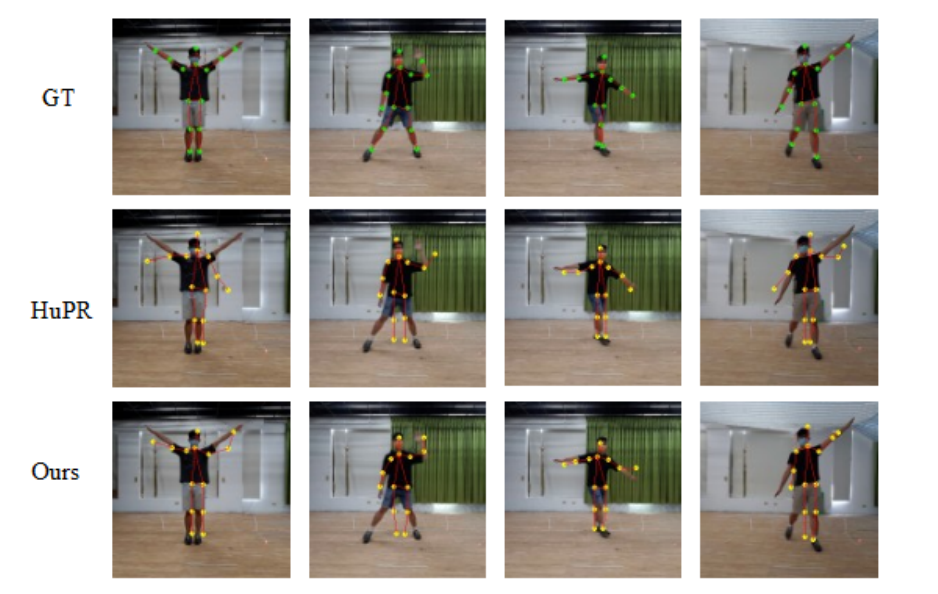}
\caption{Visualisation of predicted keypoints in inaccurate condition.}
\label{fig4}
\end{figure}

This section introduces the implementation and detailed performance of our proposed method. Ablation experiments are also conducted to compare the prediction performance of incomplete models.

\subsection{Dataset and Evaluation}

The HuPR dataset was chosen for our experiments \cite{Hupr}. 
This dataset acquires data from two identical radars. 
One radar sensor is rotated $90^\circ$ in the antenna plane with respect to the other, with one radar focusing on the horizontal plane and the other on the vertical plane. 
Unlike some radar datasets that provide processed data, e.g., point clouds \cite{Mars,milipoint,Radhar}, heatmaps \cite{RPM}, etc., which lose the purity of radar data, HuPR provides raw radar ADC signals. 
Table \ref{tab:dataset} demonstrates a comparison of the data formats provided by mainstream accessible mmWave radar-based human posture datasets. 
The HuPR dataset is the only dataset that provides radar signals that have not been manually processed. 
This allows for more possibilities for our data processing and feature extraction methods.


\begin{table}[tb]
    \centering
    \arrayrulecolor{black}
    \renewcommand{\arraystretch}{1} 
    \caption{ Data Formats for MmWave Radar Human Pose Datasets}
    \large
    \resizebox{\linewidth}{!}{
    \begin{tabular}{cccccc}
        \toprule
         \textbf{$Dataset$} &  \textbf{$Radar$}     &  \textbf{$Raw Signal$} &  \textbf{$Point Cloud$} &  \textbf{$Heatmap$}    \\ \midrule 
        \hline
        \renewcommand{\arraystretch}{1}
         Mars\cite{Mars}  &  TI IWR1443        & { \Large \(\circ\)} & { \Large \(\bullet\)} & { \Large \(\circ\)}\\
        mRI\cite{mri}  &  TI IWR1443        & { \Large \(\circ\)} & { \Large \(\bullet\)} & { \Large \(\circ\)}\\
         mmPose\cite{mmPose}  &  TI AWR1642        & { \Large \(\circ\)} & { \Large \(\bullet\)} & { \Large \(\circ\)}\\
         miliPoint\cite{milipoint}  &  TI IWR1843        & { \Large \(\circ\)} & { \Large \(\bullet\)} & { \Large \(\circ\)} \\
         \textbf{HuPR}\cite{Hupr}  &  TI IWR1843        & { \Large \(\bullet\)} & { \Large \(\circ\)} & { \Large \(\circ\)} \\
         mmPose-FK\cite{mmpose-fk} &  TI IWR6843        & { \Large \(\circ\)} & { \Large \(\bullet\)} & { \Large \(\circ\)}  \\
         mmBody\cite{mmbody}  &  Arbe Robotics Phoeniex        & { \Large \(\circ\)} & { \Large \(\bullet\)} & { \Large \(\circ\)}  \\
         HIBER\cite{RFmask} &  TI MMWCAS-RF-EVM        & { \Large \(\circ\)} & { \Large \(\circ\)} & { \Large \(\bullet\)}  \\
        
        \bottomrule
        \hline
    \end{tabular}
    }
    \label{tab:dataset}
\end{table}

The HuPR dataset includes data from 235 acquisition sequences in an indoor environment. 
Each sequence contains RGB camera frames, horizontal radar frames, and vertical radar frames that are one minute in duration. 
The two radars and the camera are synchronized and configured to capture 10 frames per second (FPS), so each sequence has a set of 600 camera-radar-radar frames. 
In each sequence, one person performs a static action, a standing hand wave, and a walking hand wave. 
These poses imitate and demonstrate the basic movements of a person in a medical care scenario, such as performing arm and walking rehabilitation therapy.
The dataset also provides ground truth generated by the human pose estimation network HRNet \cite{HRnet} from RGB frames.


In alignment with established norms in the field, our evaluation framework for 2D keypoints employs average precision (AP) metrics calculated across various levels of object keypoint similarity (OKS) \cite{MicrosoftCoco}. 
The OKS is calculated by
\begin{equation}
\begin{aligned}
\mathrm{OKS}_{p}=\frac{\sum_i \exp (-d_{p_i}^2 / 2 S_p^2 \sigma_i^2) \delta\left(v_{p_i}=1\right)}{\sum_i \delta\left(v_{p_i}=1\right)},
\end{aligned}
\end{equation}
%
%
%
where \( p_i \) denotes the ID of the targeted keypoint. 
\(d_{p_i}\) represents the Euclidean distance between the predicted keypoint and the ground truth keypoint for the \(i\)th keypoint on person \(p\). 
\( S_p \) is the scale of the person based on its area, calculated from the ground truth box of the person. 
\( \sigma_i \) is the normalization factor of the corresponding skeletal point, reflecting the influence of the current skeletal point on the whole.  
\( \delta\left(v_{p_i}=1\right) \) indicates that the predicted keypoint \( p_i \) is visible within the observation range.


This allows the evaluation of model accuracy in detecting 2D keypoints across 14 critical human body positions, including head, neck, shoulders, elbows, wrists, hips, knees, and ankles.
To offer a nuanced understanding of model performance, three distinct AP metrics are used: \(AP^{50}\), \(AP^{75}\), and \(AP\). 
These metrics represent varying degrees of OKS stringency, with \(AP^{50}\) and \(AP^{75}\) indicating more lenient and stringent OKS constraints, respectively. 
The metric $AP$ calculates the mean average precision across a range of 10 OKS thresholds, specifically at 0.5, 0.55, and incrementally up to 0.95, providing a comprehensive overview of the model's performance across a broad spectrum of precision requirements.

\subsection{Implementation Details}

We take 193 sequences from the HuPR dataset for training, 21 for validation, and 21 for testing. The sequences are chosen as HuPR network, which is considered as baseline, for fair comparison of results.

ProbRadarM3F is implemented using Python and Pytorch. The training of our network is processed on a single NVIDIA V100 GPU. The network is trained using the step decay learning rate strategy, with initial learning rate $0.00005$ and reduction $0.999$ times per $2000$ iterations. The Adam is employed as optimizer, and the batch size is set to $32$. Some other parameters in the experiment are set as follows. In CFAR processing, the settings of the guard unit and the reference unit are set to 5 and 16.
The depth of positional encoding for the probability map is set to 32.
Table \ref{tab:parameters} shows how the model performs for some of the keypoints and the overall prediction for different positional encoding layers.
The number of input frames is 8 for joint processing of multi-frame data. 
The convolution network contains basic blocks from ResNet \cite{Resnet}. The size of convolution kernel for 3D convolutional processing is \(3\times3\times3 \) and the ReLU activation function is used. These parameters are selected based on optimal performance in experiments.

\begin{table}[h]
  \centering
  \arrayrulecolor{black}
  \renewcommand{\arraystretch}{1.5} 
  \caption{Comparison of Different Positional Encoding Layers}
  \begin{tabular}{l|c|c|c|c}
    \hline
    \multirow{2}{*}{\textbf{$Layers$} }&\multicolumn{4}{c}{{AP}} \\
     & \multicolumn{1}{l}{$Head$} & \multicolumn{1}{l}{\textbf{$Elbow$}} & \multicolumn{1}{l}{\textbf{$Wrist$}}  & \multicolumn{1}{l}{\textbf{$Total$}} \\
    \hline
    8 & \multicolumn{1}{l}{80.9} & \multicolumn{1}{l}{48.8} & \multicolumn{1}{l}{27.0}& \multicolumn{1}{l}{69.2} \\ 
    
    16 & \multicolumn{1}{l}{82.9} & \multicolumn{1}{l}{50.9} & \multicolumn{1}{l}{26.9} & \multicolumn{1}{l}{68.3} \\ 
    
   \textbf{32} & \multicolumn{1}{l}{\textbf{81.1}} & \multicolumn{1}{l}{\textbf{52.2}} & \multicolumn{1}{l}{\textbf{28.2}} & \multicolumn{1}{l}{\textbf{69.9}} \\

    64 & \multicolumn{1}{l}{80.0} & \multicolumn{1}{l}{45.8} & \multicolumn{1}{l}{23.5}
    & \multicolumn{1}{l}{0.65}\\
    \hline
  \end{tabular}
  \label{tab:parameters}
\end{table}

\begin{table*}[t!]
  \centering
  \arrayrulecolor{black}
  \renewcommand{\arraystretch}{1.5}
  \caption{Comparison of keypoint Accurate Precison}
    \begin{tabular}{l|cccccccc}
    \hline
          & \multicolumn{8}{c}{$AP$} \\
    \textbf{Model} & \multicolumn{1}{l}{Head} & \multicolumn{1}{l}{Neck   } & \multicolumn{1}{l}{Shoulder} & \multicolumn{1}{l}{Elbow} & \multicolumn{1}{l}{Wrist} & \multicolumn{1}{l}{ Hip } & \multicolumn{1}{l}{Knee} & \multicolumn{1}{l}{ Ankle} \\
    \hline
    RF-Pose \cite{ThroughWallHPE}  & 61.0  & 65.3  & 52.5  & 16.1  & 6.3  & 73.5  & 65.7  & 62.0\\
    HuPR \cite{Hupr}  & 77.5  & 81.9  & 70.3  & 45.5  & 22.3  & 88.1  & 82.2  & 73.1 \\
    Ours  & \textbf{81.1}  & \textbf{83.6}  & \textbf{78.1}  & \textbf{52.2}  & \textbf{28.2}  & \textbf{92.9}  & \textbf{88.1}  & \textbf{75.8} \\
   
    \hline
    \end{tabular}
  \label{tab:1}
\end{table*}

\begin{table}[!ht]
  \centering
  \arrayrulecolor{black}
  \renewcommand{\arraystretch}{1.5} 
  \caption{Comparison of State-of-the-art Approaches and Our Proposed Model}
  \begin{tabular}{l|c|c|c}
    \hline
    \textbf{Model} & \multicolumn{1}{l}{$AP$} & \multicolumn{1}{l}{\textbf{$AP^{50}$}} & \multicolumn{1}{l}{\textbf{$AP^{75}$}} \\
    \hline

    RF-Pose \cite{ThroughWallHPE} & \multicolumn{1}{l}{41.4} & \multicolumn{1}{l}{82.9} & \multicolumn{1}{l}{37.0} \\ 
    
    HuPR \cite{Hupr} & \multicolumn{1}{l}{63.4} & \multicolumn{1}{l}{97.0} & \multicolumn{1}{l}{74.0} \\ 
    
    Ours & \multicolumn{1}{l}{\textbf{69.9}} & \multicolumn{1}{l}{\textbf{98.5}} & \multicolumn{1}{l}{\textbf{86.9}} \\
    \hline
  \end{tabular}
  \label{tab:2}
\end{table}

\subsection{Results and Analysis}


Experiments on the HuPR dataset were performed to evaluate the effectiveness of the proposed strategies. 
For fair comparison, we compared with the baseline RF-Pose\cite{ThroughWallHPE} and the state-of-the-art method HuPR\cite{Hupr}. There are a few earlier studies and most of them neither publish their hardware configurations nor share their training data, so it is challenging to reproduce their results.
Table \ref{tab:2} compares the average precision values of our proposed methods with RF-Pose and the SOTA method.
ProbRadarM3F shows advantages in AP at various precision levels, illustrating that our method outperforms RF-Pose on the same dataset.
Compared to the state-of-the-art method, our method achieves higher scores on every metric, especially in $AP$ and \(AP^{75}\), with gains of 6.5\(\%\) and 12.9\(\%\), respectively. 
The increase in \(AP^{75}\) is a stringent indicator of our method's ability to accurately predict human keypoints.


The specific precision of each keypoint is detailed in Table \ref{tab:1}. 
Our method has improved precision at every joint keypoint compared to the state-of-the-art. 
The best performance is observed in the hip joints, where $AP$ reaches 92.9\(\%\). 
The greatest improvement is seen in the shoulder joints, with an increase $AP$ of 8.2\(\%\). 
Reflected signals from the torso joints are strong and contain significant information with little noise, resulting in more accurate predictions. 
However, estimating arm and wrist joints remains challenging due to richer and finer arm movements and the radar's limitations in capturing information from small reflective surfaces away from the torso. 
Despite this, our work shows accuracy gains of 6.7\(\%\) in the elbow and 5.9\(\%\) in the wrist.


Fig. \ref{fig3} illustrates the performance of our model under different types of actions. 
GT represents the ground truth keypoints generated from HRNet \cite{HRnet} that precisely follow the actions in RGB frames, while HuPR and Ours are the predictions of the baseline and our ProbRadarM3F for the same frame, respectively. 
As shown, the predicted keypoints mostly align with the ground truth. 
To present realistic results, some frames where the prediction is not accurate enough are displayed in Fig. \ref{fig4}. 
It can be seen that our model loses accuracy when the target moves quickly. 
Although it predicts the torso and head locations well, there is still room for improvement in the wrist and foot joints. 
This observation highlights a critical area for future improvement, particularly in enhancing the model's sensitivity to high-motion extremities.


Therefore, by analyzing both the overall and individual results, our work effectively improves the precision of mmWave radar-based human skeletal pose estimation. 
It demonstrates that the probability map-guided positional encoding method effectively mines information from radar signals that might have been previously overlooked.

\subsection{Ablation Study}


A series of ablation studies were conducted to evaluate the individual contributions of various components within our proposed framework, ProbRadarM3F. 
These studies were performed on the HuPR dataset under consistent training, validation, and testing set settings.

\begin{itemize} 
\item[$\bullet$]
To determine the impact of the ProbPE branch, the network was operated with only the FFT branch, excluding the integration of features from the positional encoding guided by probability maps. 
The experimental results are shown in Table \ref{tab:3}. 
\begin{table}[ht!]
  \centering
  \renewcommand{\arraystretch}{1.5} 
  \caption{Ablation Study}
    \begin{tabular}{l|c|c|c}
    \hline
    \textbf{Model} & \multicolumn{1}{l}{$AP$} & \multicolumn{1}{l}{\textbf{$AP^{50}$}} & \multicolumn{1}{l}{\textbf{$AP^{75}$}} \\
    \hline
    
    Ours (without ProbPE Branch) & \multicolumn{1}{l}{66.1} & \multicolumn{1}{l}{97.1} & \multicolumn{1}{l}{79.2} \\
    
    Ours (without Probability Maps Generation) & \multicolumn{1}{l}{67.6} & \multicolumn{1}{l}{98.4} & \multicolumn{1}{l}{84.2} \\

    Ours (Complete ProbRadarM3F Model) & \multicolumn{1}{l}{\textbf{69.9}} & \multicolumn{1}{l}{\textbf{98.5}} & \multicolumn{1}{l}{\textbf{86.9}} \\
    
    \hline
  \end{tabular}
  \label{tab:3}
\end{table}
In the absence of the ProbPE branch, the model achieved lower precision in keypoint prediction compared to the complete ProbRadarM3F model, particularly under stringent OKS constraints. 
Specifically, excluding the ProbPE branch resulted in a decrease of 3.8\(\%\) in $AP$ and a more pronounced 7.7\(\%\) reduction in $AP^{75})$. 
The importance of the ProbPE branch is evident, highlighting the significant role of position information in the model's prediction performance. 

\item[$\bullet$] 
In addition to isolating the effect of the ProbPE branch as a whole, the influence of probability maps in the ProbPE branch was specifically assessed. 
The complete model conducts positional encoding guided by probability maps, which are pivotal in generating refined positional features. 
To quantify the contribution of these probability maps, an ablation experiment was conducted by excluding the probability map generation step and directly applying positional encoding to the data after DOA. 
As shown in Table \ref{tab:3}, precision improved compared to using the FFT branch independently.
However, the absence of probability maps resulted in a noticeable decrease in performance, leading to a reduction of $2.3\%$ in $AP$ when compared to the complete model.
The results demonstrate the importance of probability maps. 
Without the guidance of probability maps, positional encoding becomes less effective. 
Consequently, the probability maps are not merely auxiliary but integral to ensuring that the positional encoding maximally captures and utilizes the spatial context inherent in the radar signals. 
The probability maps enhance the model's ability to accurately determine the probability of target presence at specific azimuth and elevation coordinates, which is crucial for accurately encoding positional information. 
\end{itemize}

\section{CONCLUSION}\label{sec conclusion}

In this paper, mmWave radar was applied to estimate human skeletal poses. 
This is an important component of smart sensor systems used in medical care IoT applications, particularly in addressing privacy concerns. 
ProbRadarM3F is designed to apply millimeter wave radar for human skeletal pose estimation. 
This model significantly improves the extraction and use of hidden information from radar signals, by introducing the ProbPE branch, which generates probability maps based on estimating the likelihood of a specific position. 
These probability maps are then used to effectively extract position features using a positional encoding method. 
Additionally, we combined positional features with features obtained from the FFT branch to enhance the human body keypoint features produced by the model. 
Our experiments on the HuPR dataset demonstrated that ProbRadarM3F outperforms existing methods, indicating the effectiveness of our approach. 
Our results highlight that there is valuable information, such as positional data, present in the radar heatmap or even in the radar signal itself, which should not be considered redundant in human pose estimation tasks.\vspace*{-.5pc}

\balance

\small
\bibliographystyle{IEEEtran}
\bibliography{ref}

\end{document}